\pdfoutput=1

\documentclass[11pt]{article}

\usepackage[]{emnlp2021}

\usepackage{times}
\usepackage{latexsym}

\usepackage[T1]{fontenc}

\usepackage[utf8]{inputenc}

\usepackage{microtype}

\usepackage{graphicx}
\usepackage{multirow}
\usepackage{microtype}
\usepackage{multicol}
\usepackage{footmisc}
\usepackage{ctable}
\usepackage{amsmath}
\usepackage{newtxtext,newtxmath}
\usepackage[labelfont=bf]{caption}
\usepackage{subcaption}
\usepackage{mathrsfs}
\usepackage{adjustbox}
\usepackage{bbding}
\usepackage{pifont}
\usepackage{soul}
\newcommand{\cmark}{\ding{51}}%
\newcommand{\xmark}{\ding{55}}%
\colorlet{soulorange}{orange!30}

\usepackage{xcolor}
\definecolor{my_red}{HTML}{C93429}
\definecolor{my_blue}{HTML}{007FFF}

\title{Vision Guided Generative Pre-trained Language Models for \\ Multimodal Abstractive Summarization}

\author{Tiezheng Yu\thanks{$^*$ The two authors contribute equally.} \,, Wenliang Dai$^*$, Zihan Liu, Pascale Fung \\
Center for Artificial Intelligence Research (CAiRE)\\
Department of Electronic and Computer Engineering\\
The Hong Kong University of Science and Technology, Clear Water Bay, Hong Kong\\
\texttt{\{tyuah,wdaiai,zliucr\}@connect.ust.hk},  \texttt{pascale@ece.ust.hk}}

\begin{document}
\maketitle
\begin{abstract}

Multimodal abstractive summarization (MAS) models that summarize videos (vision modality) and their corresponding transcripts (text modality) are able to extract the essential information from massive multimodal data on the Internet. Recently, large-scale generative pre-trained language models (GPLMs) have been shown to be effective in text generation tasks. However, existing MAS models cannot leverage GPLMs' powerful generation ability. To fill this research gap, we aim to study two research questions: 1) how to inject visual information into GPLMs without hurting their generation ability; and 2) where is the optimal place in GPLMs to inject the visual information?
In this paper, we present a simple yet effective method to construct vision guided (VG) GPLMs for the MAS task using attention-based add-on layers to incorporate visual information while maintaining their original text generation ability. 
Results show that our best model significantly surpasses the prior state-of-the-art model by 5.7 ROUGE-1, 5.3 ROUGE-2, and 5.1 ROUGE-L scores on the How2 dataset~\citep{sanabria2018how2}, and our visual guidance method contributes 83.6\% of the overall improvement. Furthermore, we conduct thorough ablation studies to analyze the effectiveness of various modality fusion methods and fusion locations. 
\footnote{The code is available at: \url{https://github.com/HLTCHKUST/VG-GPLMs}}

\end{abstract}

\section{Introduction} \label{sec:intro}

\begin{figure}[t]
    \centering
    \includegraphics[width=\linewidth]{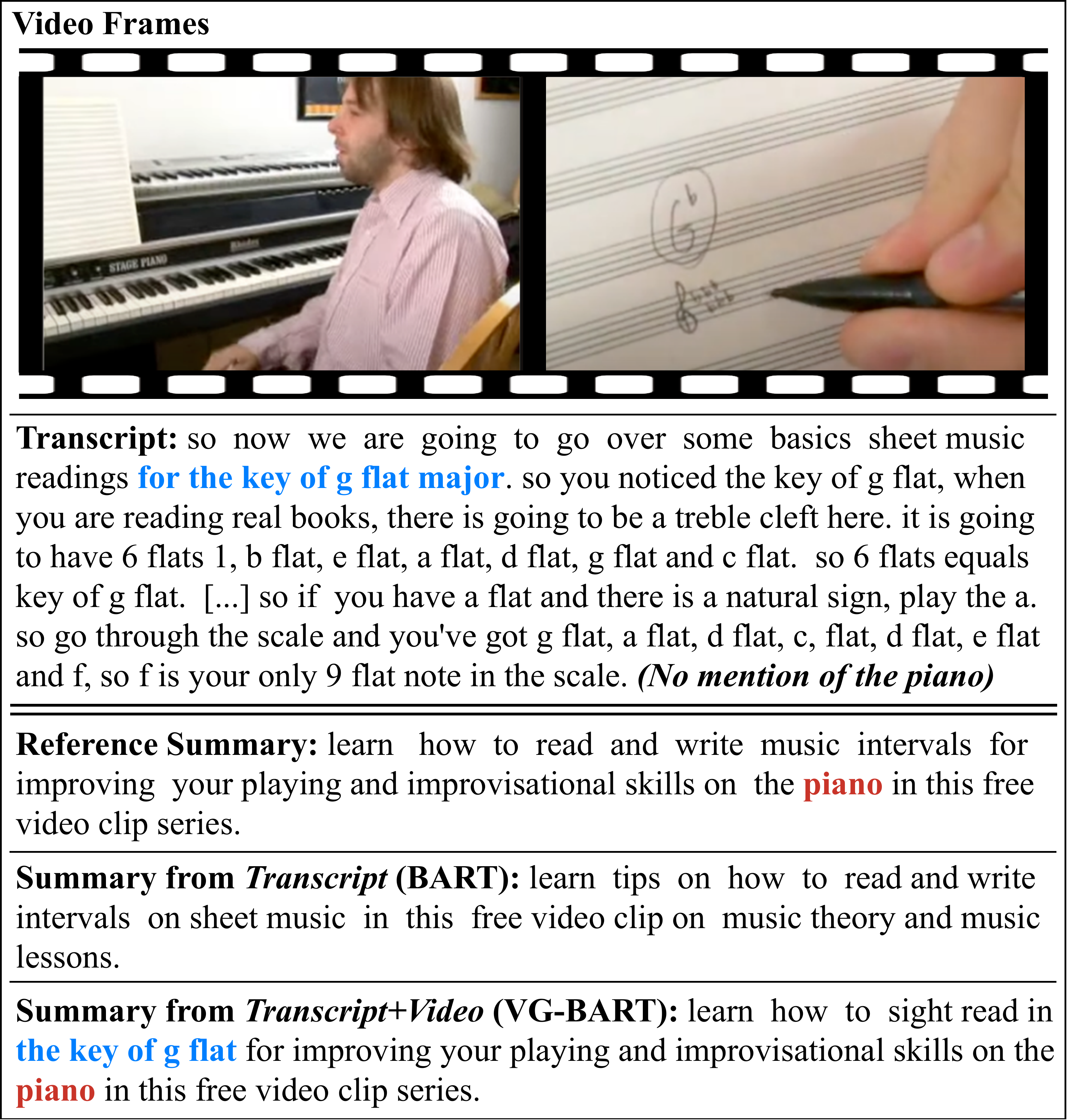}
    \caption{An example of MAS. As input data, we show two representative video frames and the transcript, with [...] representing omitted unimportant text. As illustrated, some information is emphasized (e.g. \textbf{\textcolor{my_blue}{the key of g flat}}) or only exists (e.g. \textbf{\textcolor{my_red}{piano}}) in the visual signal. We also compare the human-generated reference summary and our model-generated summaries with/without video frames in the input data.}
    \label{fig:intorduce_task}
\end{figure}

Multimodal abstractive summarization (MAS) aims to take advantage of data from multiple modalities and provides a short, concise and readable textual summary to let users quickly acquire their essential information~\cite{sanabria2018how2,palaskar2019multimodal,liu2020multistage}. MAS has become an increasingly popular research area thanks to the proliferation of online multimedia content and the increasing availability of multimodal data. 

As illustrated in Figure~\ref{fig:intorduce_task}, the MAS models need to generate a concise summary by effectively utilizing two modalities: a video and its transcript. Therefore, we emphasize that leveraging a powerful text generation model and an effective combination of the vision and text modalities are key to constructing good MAS models.
Recently, Transformer-based~\citep{vaswani2017attention} sequence-to-sequence (Seq2Seq) large-scale generative pre-trained language models (GPLMs), such as BART~\citep{lewis2019bart}, T5~\citep{raffel2019exploring}, PEGASUS~\citep{pmlr-v119-zhang20ae} and ProphetNet~\citep{qi-etal-2020-prophetnet}, have shown remarkable performance on text generation tasks, including abstractive text summarization.
However, leveraging and adapting GPLMs to MAS is still an unexplored research direction. To explore this direction, two main questions need to be answered: Firstly, \textit{how} can we inject visual information into the text-only GPLMs so that the models can understand both modalities and allow cross-modal interactions, and more importantly, how can this injection operation be conducted without damaging GPLMs' original text generation ability? Secondly, \textit{where} is the optimal place in GPLMs to inject the visual information? This needs to be explored, as there are many sub-layers in the encoder and decoder of GPLMs and a sub-optimal location might result in unsatisfactory performance.

In this paper, to fill the research gap, we present a simple yet very effective method to construct vision guided (VG) GPLMs (VG-BART and VG-T5) for the MAS task.
Specifically, to answer the first of the aforementioned questions, we insert attention-based add-on layers to GPLMs to incorporate visual information without modifying the original architecture. In this way, all the pre-trained model weights can be used during fine-tuning so as to preserve their original text generation ability. We try with two types of attention mechanisms for the text-vision fusion and interaction: 1) Cross-modal Dot-product Attention; and 2) Cross-modal Multi-head Attention. Moreover, we also investigate the effects of using a forget gate and a visual transformer encoder along with the attention mechanisms. To answer the second question, we enumerate almost all possible locations in GPLMs for injecting add-on layers, and show a thorough comparison and analysis in Section~\ref{sec:results_and_analysis}. We evaluate our models on the How2 dataset~\citep{sanabria2018how2}. Experimental results demonstrate that our best model surpasses the prior state-of-the-art model by 5.7 ROUGE-1, 5.3 ROUGE-2, and 5.1 ROUGE-L scores. To ensure this improvement does not purely come from the GPLMs, we also evaluate the corresponding text-only model, and the results show that the injected visual guidance contributes 83.6\% of the overall improvement on average of all ROUGE scores.

Our contributions in this work are threefold:
\begin{itemize}
    \item To the best of our knowledge, we are the first to inject visual information into text-only GPLMs, and to use it for the MAS task.
    \item We systematically study two research questions: 1) how to inject visual information into GPLMs without hurting their generation ability; and 2) where is the optimal place in GPLMs to inject the visual information?
    \item Our model significantly outperforms the state-of-the-art model on the How2 dataset, and the injected visual guidance contributes 83.6\% of the overall improvement. 
\end{itemize}

\section{Related Work}
\subsection{Abstractive Text Summarization}
Abstractive text summarization aims to generate short, concise and readable text that can capture the most salient information of the input documents. Thanks to the Seq2Seq framework \citep{sutskever2014sequence} and attention mechanisms, deep neural networks have achieved remarkable results on summarization tasks~\citep{paulus2017deep,zhang2020pegasus,yu2021adaptsum}. Recently, GPLMs~\citep{lewis2019bart, raffel2019exploring, pmlr-v119-zhang20ae, qi-etal-2020-prophetnet} have been widely used in abstractive text summarization and have achieved start-of-the-art performance. 
The most significant difference between abstractive text summarization and multimodal abstractive summarization lies in whether the input contains data of more than one modality.

\subsection{Multimodal Abstractive Summarization}
Recently, many studies have been performed on multimodal learning~\citep{mroueh2015deep, antol2015vqa, donahue2015long, zadeh2017tensor,dai-etal-2020-modality,dai-etal-2021-multimodal}. However, only a few have investigated MAS. \citet{li2017multi} collected a multimodal corpus of news articles containing 500 videos of English news articles paired with human-annotated summaries. \citet{sanabria2018how2} introduced the How2 dataset, which contains about 2,000 hours of short instructional videos, each coming with a summary of two to three sentences. \citet{palaskar2019multimodal} proposed a multi-source Seq2Seq model with hierarchical attention to integrate information from different modalities into a coherent summary. Meanwhile, \citet{liu2020multistage} proposed a multi-stage fusion network with the fusion forget gate module, which can model the fine-grained interactions between multi-source modalities. To the best of our knowledge, no previous work has leveraged GPLMs' generation ability to tackle the MAS task, and we are the first to systematically study multiple multimodal fusion methods based on GPLMs.

\begin{figure*}[t]
    \centering
    \includegraphics[width=\linewidth]{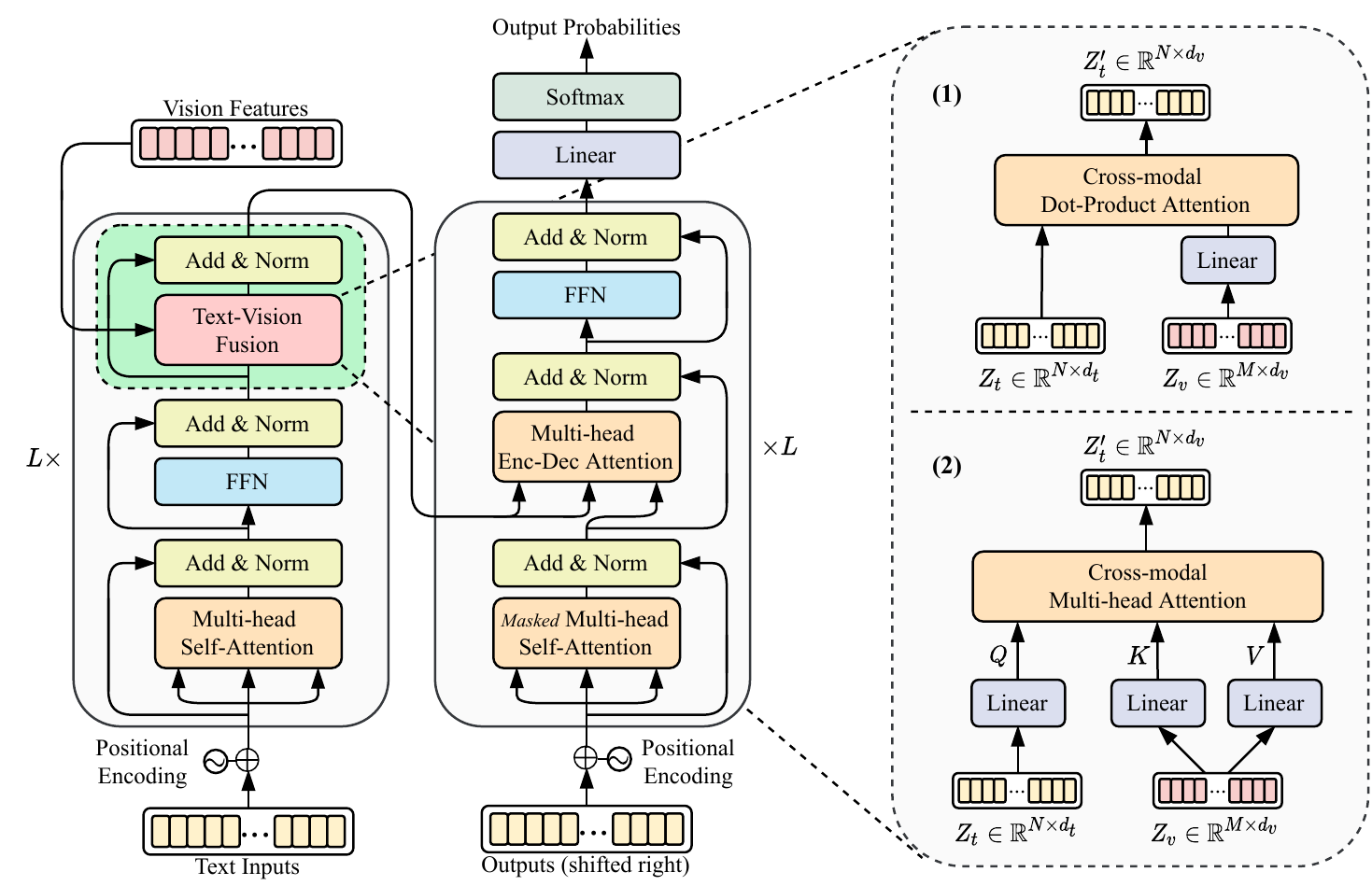}
    \caption{An overview of our proposed VG GPLMs. It is built based on the Transformer-based Seq2Seq GPLMs (\textit{left}). To inject visual information, we insert add-on sub-layers (\textit{the green dashed block}) by mainly leveraging two kinds of attention-based text-vision fusion mechanism (\textit{right}): 1) Cross-modal Dot-Product Attention; and 2) Cross-modal Multi-head Attention. Although we draw the add-on sub-layers in the encoder, they can also be placed in the decoder in a similar way. We compare the effects of different injection locations in Section~\ref{sec:results_and_analysis}.}
    \label{fig:main_model}
\end{figure*}

\subsection{Vision-Language Large Pre-trained Transformer Models}
With the remarkable success of large-scale unsupervised pre-training in NLP~\citep{devlin2019bert,liu2019roberta,radford2019language}, pre-training large \textbf{v}ision-\textbf{l}anguage (VL) models has also become more and more popular in recent years. Rather than designing task-specific architectures, pre-training results in a general backbone model by feeding it with a large amount of data and then fine-tune it to different downstream tasks. Among the current VL pre-training work, most has been focusing on VL understanding by training BERT-style Transformer models~\citep{Sun2019VideoBERTAJ,tan-bansal-2019-lxmert,Su2020VL-BERT,Li2020UnicoderVLAU,chen2020uniter} and finetune them on various VL classification tasks~\cite{balanced_vqa_v2,zellers2019vcr,suhr-etal-2019-corpus}. These models usually receive a pair of text and image as input, where the image is processed into  objects~\citep{zhang2021vinvl}, patches~\citep{pmlr-v139-kim21k}, or pixels~\citep{Huang2020PixelBERTAI} before feeding into the VL model. For VL text generation, \citet{Zhou2020UnifiedVP} presented a model for both visual question answering and image captioning~\citep{capeval2015}. Additionally, \citet{Cho2021UnifyingVT} introduced an encoder-decoder Transformer model that unifies all VL tasks as generative tasks. 
Although prior work has made much progress on VL pre-training, the problem of generating text given text and video input (E.g. the How2 dataset) is not well studied under the VL pre-training setting, except by \citet{Luo2020UniViLMAU}, who proposed a dual-stream model for both VL classification and generation with video data. However, compared to GPLMs in NLP such as BART~\citep{lewis2019bart} and T5~\citep{raffel2019exploring}, their text generation ability is limited as the training data is much smaller.

In this paper, we propose to tackle VL tasks and utilize the advantage of pre-training from a different angle by inserting add-on layers to the text-only GPLMs and fine-tuning them on multimodal tasks to incorporate visual information. This takes advantage of GPLMs' superior generation ability to generate vision-aware texts. Of the very few works that have also considered this direction, \citet{rahman-etal-2020-integrating} proposed the multimodal adaptation gate, which fuses data of other modalities to the textual embeddings in BERT. However, their method requires all modalities to have the same sequence length, which is rare for most datasets. Additionally, they only attempted to address the sentiment analysis task and did not explore text generation.



\section{Vision Guided GPLMs} \label{sec:VG_LM}
To take advantage of the superior text generation ability of the text-only Seq2seq GPLMs and adapt them to the MAS task, we present Vision guided (VG) GPLMs. Specifically, we leverage BART~\citep{lewis2019bart} and T5~\citep{raffel2019exploring} to construct VG-BART and VG-T5.


In this section, we start by revisiting the text-only Seq2seq GPLMs in Section~\ref{sec:overview_lm}. These serve as the backbone of our proposed model and also one of the baselines. Then, we discuss the approach for extracting visual features from video clips in Section~\ref{sec:vision_features}, as well as how to further process them. Finally, in Section~\ref{sec:fusions}, we introduce two types of text-vision fusion mechanism to guide the GPLMs to generate vision-aware summaries.

\subsection{Overview of GPLMs for Summarization} \label{sec:overview_lm}
Transformer-based~\citep{vaswani2017attention} Seq2Seq GPLMs generalize architectures like BERT~\citep{devlin2019bert} and GPT~\citep{radford2018improving} by including a bi-directional encoder and a uni-directional (left-to-right) decoder. 
An overview of this architecture is depicted on the left side of Figure~\ref{fig:main_model} (except the green dashed block).

At the entry of the GPLM, the input text is first tokenized and converted to a sequence of token embeddings $X_t \in \mathbb{R}^{N \times d_t}$, in which $N$ is the sequence length and $d_t$ is the feature dimension. To retain the positional information, positional encodings~\citep{NIPS2017_3f5ee243} $E_{pe} \in \mathbb{R}^{N \times d_t}$ are added to the token embeddings pointwisely (Eq.~\ref{eq:emb_pe}), which forms the input features $Z_0^{enc}$ to the encoder.
\begin{align}
    Z_0^{enc} &= X_t + E_{pe} \label{eq:emb_pe}
\end{align}
As illustrated in Figure~\ref{fig:main_model}, the encoder is composed of a stack of $L$ encoder layers, each containing two sub-layers: 1) Multi-head Self-Attention (MSA, Eq.~\ref{eq:msa_enc}) and 2) Feed-Forward Network (FFN, Eq.~\ref{eq:ffn_enc}). In addition, after each sub-layer, there is a residual connection~\citep{He2015,wang-etal-2019-learning-deep} followed by a layer normalization (LN)~\citep{Ba2016LayerN}. See Appendix \ref{appendix:msa} and \ref{appendix:ffn} for more details of the MSA and FFN.
\begin{align}
    Z_l^{enc\prime} &= \text{LN}(\text{MSA}(Z_{l-1}^{enc}) + Z_{l-1}^{enc}) \label{eq:msa_enc}\\
    Z_l^{enc} &= \text{LN}(\text{FFN}(Z_l^{enc\prime}) + Z_l^{enc\prime}) \label{eq:ffn_enc}
\end{align}

Similar to the encoder, the decoder also consists of a stack of $L$ decoder layers, but with two differences. Firstly, the MSA is masked to prevent positions from attending to subsequent positions (keep the decoder in a left-to-right direction). Secondly, there is one more multi-head encoder-decoder attention sub-layer, which uses the decoder embeddings to attend over the output embeddings of the encoder to incorporate the encoded information.

Specifically, in our experiments, we adopt the pre-trained BART~\citep{lewis2019bart} and T5~\citep{raffel2019exploring}, which both follow this architecture with different training schemes. To fine-tune them on the abstractive text summarization task, the input to the encoder is the article or transcript, and the decoder learns to generate the summaries. 

\subsection{Video Feature Extraction} \label{sec:vision_features}
For each video clip, following previous works~\citep{sanabria2018how2,palaskar2019multimodal,khullar-arora-2020-mast}, a 2048-dimensional feature representation is extracted for every 16 non-overlapping frames using a 3D ResNeXt-101 model~\citep{Hara_2018_CVPR}, which is pre-trained on the Kinetics dataset~\citep{Kay2017TheKH}. Therefore, each data sample will have a sequence of 2048-$d$ vision feature vectors of length $M$. These features can be used directly as the visual input to the text-vision fusion mechanism. 

In addition, in order to better model the intra-modal dynamics and enhance the vision specific temporal information, we further process the extracted sequence of visual features using a Transformer~\citep{NIPS2017_3f5ee243} encoder (VTF) with positional encodings. 
Experiments illustrate that this additional encoding process can further boost the performance of our model (Section~\ref{sec:results_and_analysis}).

\subsection{Text-vision Fusion} \label{sec:fusions}
As exhibited in Figure~\ref{fig:main_model}, we insert a third sub-layer (the green dashed block) into each encoder layer, which contains the text-vision fusion mechanism and also a residual connection followed by a layer normalization. We propose two types of text-vision fusion mechanism, as shown on the right-hand side of the figure. Given the textual input $Z_t \in \mathbb{R}^{N \times d_t}$ and visual input $Z_v \in \mathbb{R}^{M \times d_t}$, the fusion mechanism produces vision guided output $Z_t' \in \mathbb{R}^{N \times d_t}$ that has a same dimension as the textual input, which allows the continual stacking of layers. 

\paragraph{Dot-product Attention Based Fusion.} Before performing dot-product attention between the textual and visual features, we first project the visual features $Z_v$ to the same dimensional space as the textual features (Eq.~\ref{eq:dp_proj}). Then, we calculate the dot-product and apply the softmax function to get the attention score matrix $A$ (Eq.~\ref{eq:attn}). Finally, the input textual features $Z_t$ are concatenated with the attention weighted visual features $AZ_v$ and then projected by another linear transformation to output the vision guided textual features $Z_t'$ (Eq.~\ref{eq:fuse_out1}).
\begin{align}
    Z_v' &= Z_v W_1, \text{ } Z_v' \in \mathbb{R}^{M \times d_t} \label{eq:dp_proj}\\
    A &= \text{Softmax}(Z_t Z_v^{\prime T}), \text{ } A \in \mathbb{R}^{N \times M} \label{eq:attn}\\
    Z_t' &= \text{Concat}(Z_t, AZ_v) W_2 \label{eq:fuse_out1}
\end{align}
Additionally, we build a variant of this fusion, which uses the linearly transformed visual features $AZ_v'$ for the concatenation in Eq.~\ref{eq:fuse_out1} instead of the original $AZ_v$. A comparison of their performance is shown in Section~\ref{sec:results_and_analysis}.

\paragraph{Multi-head Attention Based Fusion.} Inspired by prior works~\citep{Yu_2019_CVPR,tsai2019MULT}, we propose a vision guided multi-head attention mechanism for the text-vision fusion. The query $Q$ is linearly projected from the input textual features, and the key $K$ and value $V$ are linearly projected from the visual features (Eq.~\ref{eq:Q} - \ref{eq:V}). Then, a cross-modal multi-head attention (CMA) is applied to get the text queried visual features $O$ (Eq.~\ref{eq:mvga}). Finally, we obtain the vision guided output $Z_t'$ by concatenating the input textual features $Z_t$ and $O$, and linearly project it to the desired dimension (Eq.~\ref{eq:fuse_out2}).
\begin{align}
    Q &= Z_t W_q, \text{ } Q \in \mathbb{R}^{N \times d_c} \label{eq:Q}\\
    K &= Z_v W_k, \text{ } K \in \mathbb{R}^{M \times d_c} \label{eq:K}\\
    V &= Z_v W_v, \text{ } V \in \mathbb{R}^{M \times d_c} \label{eq:V}\\
    O &= \text{CMA}(Q,K,V), \text{ } O \in \mathbb{R}^{N \times d_c} \label{eq:mvga}\\
    Z_t' &= \text{Concat}(Z_t, O) W_3 \label{eq:fuse_out2}
\end{align}

\noindent In addition, we also explore the effects of using a forget gate~\citep{liu2020multistage} in the text-vision fusion. 
Given the CMA output $O \in \mathbb{R}^{N \times d_c}$ in Eq.~\ref{eq:mvga}, we construct a forget gate mask $F \in \mathbb{R}^{N \times d_c}$ (Eq.~\ref{eq:fg_mask}) and do a point-wise multiplication with $O$ to output the updated $O'$ (Eq.~\ref{eq:fg_mul}).
\begin{align}
    F &= \text{Sigmoid}(\text{Concat}(O, Z_t) W_{f}) \label{eq:fg_mask}\\
    O' &= F \otimes O \label{eq:fg_mul}
\end{align}
The forget gate can potentially remove redundant and noisy information from the video features, which also helps the model to learn to discard needless visual information to retain its pre-trained text generation ability.

\section{Experimental Settings}
\subsection{How2 Dataset}


How2 \cite{sanabria2018how2} is a large-scale dataset of open-domain videos, covering 22 different topics such as cooking, exercise, yoga and music. It consists of 79,114 short instructional videos (73,993 for training, 2,965 for validation and 2,156 for testing). Each video is accompanied by a human-generated transcript and a short text summary. At the word level, the average lengths of transcripts and summaries are 291 and 33, respectively. 


\subsection{Implementation Details}

\begin{table*}[t]
\centering
\begin{adjustbox}{width={\textwidth},totalheight={\textheight},keepaspectratio}
\begin{tabular}{c|l|cccccccccc}
\toprule
\textbf{Input}                                                            & \multicolumn{1}{c|}{\textbf{Method}} & \textbf{R-1}  & \textbf{R-2}  & \textbf{R-L}  & \textbf{B-1}  & \textbf{B-2}  & \textbf{B-3}  & \textbf{B-4}  & \textbf{M}    & \textbf{C} & \textbf{CF}   \\ \midrule
\multirow{5}{*}{Transcript}                                                  & S2S*                                  & 58.6          & 40.6          & 53.8          & 55.2          & 45.6          & 39.9          & 35.8          & 27.6          & 2.35    & -        \\
                                                                             & PG*                                   & 57.2          & 39.5          & 52.8          & 55.3          & 45.6          & 39.8          & 35.7          & 26.8          & 2.13    & -        \\
                                                                             & TF*                                   & 59.0          & 41.0          & 54.3          & 56.6          & 46.7          & 40.8          & 36.6          & 27.7          & 2.30    & -        \\
                                                                             & T5                                   & 62.8          & 45.0          & 57.5          & 60.5          & 50.4          & 44.2          & 39.6          & 30.6          & 2.76    & 61.7     \\
                                                                             & BART                                 & 64.0          & 46.4          & 58.9          & 62.4          & 52.6          & 46.4          & 42.0          & 31.7          & 2.97    & 63.9     \\ \midrule \midrule
\multirow{8}{*}{\begin{tabular}[c]{@{}c@{}}Transcript\\ +Video\end{tabular}} & HA (RNN)*                             & 60.3          & 42.5          & 55.7          & 57.2          & 47.7          & 41.8          & 37.5          & 28.8          & 2.48    & -        \\
                                                                             & HA (TF)*                              & 60.2          & 43.1          & 55.9          & 58.6          & 48.3          & 43.3          & 38.1          & 28.9          & 2.51    & -        \\
                                                                             & MFFG (RNN)$^\dagger$*                 & 62.3          & 46.1          & 58.2          & 59.1          & 50.4          & 45.1          & 41.1          & 30.1          & 2.69    & -        \\
                                                                             & MFFG (TF)*                            & 61.6          & 45.1          & 57.4          & 60.0          & 50.9          & 45.3          & 41.3          & 29.9          & 2.67    & -        \\ \cmidrule{2-12} 
                                                                             & VG-T5 (Dot-product)                  & 63.0          & 44.9          & 57.6          & 60.1          & 49.8          & 43.4          & 38.8          & 30.3          & 2.74    & 61.4     \\
                                                                             & VG-T5 (Multi-head)                   & 63.3          & 45.3          & 58.0          & 60.7          & 50.8          & 44.7          & 40.2          & 31.0          & 2.86    & 62.8          \\
                                                                             & VG-BART (Dot-product)                & 66.1          & 49.3          & 61.2          & \textbf{64.5} & \textbf{55.1} & \textbf{49.2} & \textbf{44.8} & \textbf{33.2} & \textbf{3.18}     &    66.9 \\
                                                                             & VG-BART (Multi-head)                 & \textbf{66.3} & \textbf{49.4} & \textbf{61.4} & 64.1          & 54.8          & 48.9          & 44.6          & 33.1          & \textbf{3.18}     &    \textbf{67.3} \\ \bottomrule
\end{tabular}
\end{adjustbox}
\caption{Evaluation results of baselines and our proposed models on the How2 dataset. We compare the performance of using transcript only and transcript+video. The \(\dagger\) indicates the previous state-of-the-art model.  Results with * mark are taken from the previous work \cite{liu2020multistage}. We denote ROUGE, BLEU, METEOR, CIDEr and Content F1 by R, B, M, C and CF respectively.}
\label{tab:main_results}
\end{table*}

\begin{table*}[t]
\centering
\begin{adjustbox}{width={\textwidth},totalheight={\textheight},keepaspectratio}
\begin{tabular}{c|l|cccccccccc}
\toprule
\textbf{Input}                                                               & \multicolumn{1}{c|}{\textbf{Method}}      & \textbf{R-1} & \textbf{R-2} & \textbf{R-L} & \textbf{B-1} & \textbf{B-2} & \textbf{B-3} & \textbf{B-4} & \textbf{M}   & \textbf{C}   & \textbf{CF} \\ \midrule
\multirow{4}{*}{\begin{tabular}[c]{@{}c@{}}Transcript\\ +Video\end{tabular}} & \multicolumn{1}{c|}{VG-BART (Multi-head)} & 66.3         & 49.4         & 61.4         & 64.1         & 54.8         & 48.9         & 44.6         & 33.1         & 3.18         & 67.3 \\
                                                                             & \quad w/ FG                               & 67.3         & 50.7         & 62.4         & 65.0         & 55.9         & 50.1         & 45.7         & 33.8         & 3.25         & \textbf{72.5} \\
                                                                             & \quad w/ VTF                              & 67.3         & 50.9         & 62.6         & 64.9         & 56.0         & 50.1         & 45.7         & 33.7         & 3.20         & 72.1 \\
                                                                             & \quad w/ FG+VTF                           & \textbf{68.0}& \textbf{51.4}& \textbf{63.3}& \textbf{65.2}& \textbf{56.3}& \textbf{50.4}& \textbf{46.0}& \textbf{34.0}& \textbf{3.28}& 69.7 \\ \bottomrule
\end{tabular}
\end{adjustbox}
\caption{Further Evaluation of adding forget gate (FG) and visual transformer encoder (VTF) to our best model setting in Table~\ref{tab:main_results} on the How2 dataset. VG-BART+FG+VTF largely surpasses the previous state-of-the-art model.}
\label{tab:results_fg_tf}
\end{table*}


\paragraph{Data pre-processing.} We pre-process the transcripts data by truncating or padding them into sequences of 512 tokens after tokenization. For the videos, after the feature extraction as described in Section~\ref{sec:vision_features}, we also truncate or pad the sequence length to 256.

\paragraph{Hyper-parameters.} We use BART-base and T5-base as the pre-trained GPLMs to construct VG-BART and VG-T5, in which $L=6$ for both encoder and decoder. 
For the VTF mentioned in Section~\ref{sec:vision_features}, we use a 4-layer encoder with 8 attention heads and a 2048 feed-forward dimension. 
In the decoding stage, we use beam search with a beam size of 5. The decoding process will not stop until an end-of-sequence (EOS) token is emitted or the length of the generated summary reaches to 64 tokens. 
Following~\citet{lewis2019bart} and \citet{raffel2019exploring}, we use learning rates $6\mathrm{e}^{-4}$ and $3\mathrm{e}^{-5}$ to fine-tune the pre-trained parts of model weights. While for the newly added layers, we set the learning rate to $1.5\mathrm{e}^{-4}$. For all of our experiments, we use a batch size of 120.

\paragraph{Optimizer.} During training, we use the Adam optimizer~\citep{Kingma2015AdamAM} with $\beta_1=0.9$, $\beta_2=0.999$ and a weight decay of $1\mathrm{e}^{-5}$. Additionally, we apply a scheduler to decay the learning rate to 95\% of the current one after every 10 epochs. We train all the models for 60 epochs with an early stop of 5 using the ROUGE-2 score~\citep{xiao2019extractive} on the validation set. 

\paragraph{Software and hardware.} We use the deep learning framework PyTorch~\citep{Paszke2019PyTorchAI} to implement our code and PyTorch-Lightning\footnote{\url{https://github.com/PyTorchLightning/pytorch-lightning}} for the distributed training. We use four Nvidia GeForce RTX 2080 Ti GPUs for all of our experiment.

\subsection{Baselines}
Apart from the text-only GPLMs BART~\cite{lewis2019bart} and T5~\cite{raffel2019exploring}, we use the following baselines to compare with our proposed models, including simple models that only accept text input, as well as prior state-of-the-art models that accept text and vision modalities.

\paragraph{S2S \cite{luong2015effective}.} S2S is a standard Seq2seq model that uses RNNs for both encoder and decoder with a global attention mechanism~\cite{bahdanau2014neural}.

\paragraph{PG \cite{see2017get}.} The pointer generator (PG) network augments S2S by having a copy module to reproduce key information accurately as well as mitigating the out-of-vocabulary issue.

\paragraph{TF \cite{vaswani2017attention}.} TF is the standard Transformer-based Seq2seq model, which proposes the novel multi-head attention mechanism.



\paragraph{HA (RNN/Transformer) \cite{palaskar2019multimodal}.} A multi-source Seq2seq model with hierarchical attention (HA) \cite{libovicky2017attention} that can integrates information from different modalities into a coherent output. 

\paragraph{MFFG (RNN/Transformer) \cite{liu2020multistage}.} The multistage fusion with forget gate (MFFG) model proposes a cross fusion block with forget gate and a hierarchical fusion decoder to improve multimodal generation.

\subsection{Evaluation Metrics}
Following \cite{liu2020multistage}, we use ROUGE, BLEU, METEOR, and CIDEr to evaluate the summaries. ROUGE-\{1, 2, L\} (the standard metrics for abstractive summarization) \cite{lin2003automatic} and BLEU-\{1, 2, 3, 4\} \cite{papineni2002bleu} are used to calculate the recall and precision of n-gram overlaps, respectively, between the references and the generated summaries. MENTOR \cite{denkowski2011meteor} is used to match the word stems, synonyms and paraphrases between the reference and the generated summary. CIDEr \cite{vedantam2015cider} is an image captioning metric to compute the cosine similarity between TF-IDF weighted n-grams. 

In addition, We use Content F1 \cite{palaskar2019multimodal} to measure the F1 score of the content words of the generated summary based on a monolingual alignment. Firstly, METEOR toolkit \cite{banerjee2005meteor, denkowski2014meteor} is used to obtain the alignment between the summaries and references. Then, the function words and task-specific stop words are removed from the summaries and references. Finally, the remaining content words from the summaries and references are treated as two bags of words, and the F1 scores are calculated over the alignment. Content F1 focuses more on the content and it can avoid the increase of the ROUGE score from the stop words.

We use {\fontfamily{qcr}\selectfont nlg-eval} \footnote{\url{https://github.com/Maluuba/nlg-eval}} to compute the BLEU, MENTOR and CIDEr scores, and use {\fontfamily{qcr}\selectfont rouge} \footnote{\url{https://github.com/neural-dialogue-metrics/rouge}} to compute ROUGE scores. The implementation of Content F1 scores follows \cite{palaskar2019multimodal}.

\section{Results and Analysis}
\label{sec:results_and_analysis}

\subsection{Main Results} \label{sec:main_results}
From Table~\ref{tab:main_results}, we can see that when there is only transcript in the input data, S2S and PG reach similar scores in terms of all evaluation metrics. This could be attributed to the fact that PG tends to copy the content in the transcripts while the reference summaries in the How2 dataset have a great number of novel n-grams, which are defined to be novel with respect to the transcript. 
We also observe that TF performs better than RNN-based models. It is because TF can learn better relationships between words by multi-head attention mechanism and positional embeddings. 
Furthermore, both text-only T5 and BART outperform all the baseline models by a large gap owe to their pre-trained text generation ability. Compared to T5, BART achieves higher scores mainly because it introduces a novel pre-training objective named sentence permutation.
Sentence permutation requires the model to generate the original uncorrupted text from randomly shuffled sentences, which enhances the understanding of long text and benefits the summarization task. 
Moreover, BART is even better than all previous multimodal models trained on transcript and video. 

The visual guidance consistently boosts the performance of T5 and BART by a large step. As shown in Table~\ref{tab:results_fg_tf}, our best model \texttt{VG-BART+FG+VTF} with the cross-modal multi-head attention surpasses the previous state-of-the-art model (MFFG) by 5.7 ROUGE-1, 5.3 ROUGE-2, and 5.1 ROUGE-L scores. The visual guidance contributes 83.6\% of the overall improvement on average of all ROUGE scores. 

The results of Content F1 scores in Table \ref{tab:main_results} show similar trends with other evaluation metrics. By injecting visual information, the models can generate summaries with much richer content. Table \ref{tab:results_fg_tf} shows that both forget gate (FG) and visual transformer encoder (VTF) benefit the model's performance. However, the Content F1 score is not boosted when combining FG and VTF together, which is contradictory to all other metrics. 
We conjecture that it is because the Content F1 focuses more on the content aspect, it may have some variance compare to other metrics.


\subsection{How to Inject Visual Information} \label{sec:how_to_inject}
As illustrated in Section~\ref{sec:fusions}, we mainly adopt two text-vision fusion mechanisms to inject visual information, the cross-modal dot-product attention and multi-head attention.
As shown in Table~\ref{tab:main_results}, for the VG-BART model, these two fusion mechanisms consistently improve its performance on all metrics by a comparable margin.
However, for the VG-T5 model, the cross-modal dot-product attention based fusion does not show any improvement compared to the text-only T5, while the multi-head attention base fusion still increase its performance.
We think there are two reasons behind this phenomenon. 
Firstly, as discussed in Section~\ref{sec:main_results}, BART leverages the sentence permutation method as its pre-training objective, which increases its robustness on attention-based fusion. Secondly, multi-head attention can capture different key components in the visual information from multiple aspects, which makes it more potent than the dot-product based fusion.
Additionally, as mentioned in Section~\ref{sec:fusions}, we build a variant of the dot-product attention based fusion, which achieves 66.1 ROUGE-1, 49.3 ROUGE-2 and 61.4 ROUGE-L on VG-BART. This comparable result shows that the variant does not provide further improvement.

\begin{table}[t]
\centering
\begin{adjustbox}{width=\linewidth,totalheight={\textheight},keepaspectratio}
\begin{tabular}{c|l|ccc}
\toprule
\textbf{Input}                                                               & \multicolumn{1}{c|}{\textbf{Method}} & \textbf{R-1} & \textbf{R-2} & \textbf{R-L} \\ \midrule
\multirow{2}{*}{Transcript}                                                  & T5                                   & 62.8         & 45.0         & 57.5         \\
                                                                             & BART                                 & 64.0 & 46.4         & 58.9\\ \midrule \midrule
\multirow{4}{*}{\begin{tabular}[c]{@{}c@{}}Transcript\\ +Noise\end{tabular}} & VG-T5 (Dot-product)                  & 62.5         & 43.9         & 57.0         \\
                                                                             & VG-T5 (Multi-head)                   & 62.8         & 44.6         & 57.4         \\
                                                                             & VG-BART (Dot-product)                & 63.9         & 45.6         & 58.6         \\
                                                                             & VG-BART (Multi-head)                 & 63.9         & 46.5         & 58.7         \\ \midrule \midrule
\multirow{4}{*}{\begin{tabular}[c]{@{}c@{}}Transcript\\ +Video\end{tabular}} & VG-T5 (Dot-product)                  & 63.0         & 44.9         & 57.6         \\
                                                                             & VG-T5 (Multi-head)                   & 63.3         & 45.3         & 58.0         \\
                                                                             & VG-BART (Dot-product)                & 66.1         & 49.3         & 61.2         \\
                                                                             & VG-BART (Multi-head)                 & \textbf{66.3}& \textbf{49.4}& \textbf{61.4}\\ \bottomrule
\end{tabular}
\end{adjustbox}
\caption{Results of using uniform noise to replace the visual features.}
\label{tab:noise_data}
\end{table}

To ensure the visual features really help in the learning and our add-on layers aid the understanding of them, we conduct further experiments by replacing the visual features in the input data with random noise of the same dimension and sequence length. The noise is sampled from a uniform distribution from 0 to 3, in a similar value range of the original visual features.
As depicted in Table~\ref{tab:noise_data}, VG GPLMs with random noise as visual features achieve similar or slightly worse performance compared to the text-only GPLMs. This shows the effectiveness of our method to keep GPLMs' text generation ability.
Furthermore, compared to the dot-product attention based fusion, the multi-head fusion is better at retaining GPLMs' performance, which again demonstrates its superiority.

As mentioned in Section~\ref{sec:VG_LM}, we use a forget gate (FG) to deal with the redundancy and noisy information in the visual features. Additionally, we further encode the visual features by a visual transformer encoder (VTF). Table~\ref{tab:results_fg_tf} shows that using either FG or VTF can increase the performance of VG-BART. Jointly leveraging them boosts the performance by 1.7, 2.0, and 1.9 of ROUGE-1, ROUGE-2, and ROUGE-L, respectively. 

\begin{table}[t]
\centering
\begin{adjustbox}{width=\linewidth,totalheight={\textheight},keepaspectratio}
\begin{tabular}{ccccccccc}
\toprule
\multicolumn{6}{c}{\textbf{Encoder Layer \scriptsize{(BART-base)}}}                             & \multirow{2}{*}{\textbf{R-1}} & \multirow{2}{*}{\textbf{R-2}} & \multirow{2}{*}{\textbf{R-L}} \\ \cmidrule{1-6}
1      & 2      & 3      & 4      & 5      & 6                           &                               &                               &                               \\ \midrule
\xmark & \xmark & \xmark & \xmark & \xmark & \multicolumn{1}{c|}{\xmark} & 64.0                         & 46.4                         & 58.9                         \\  \midrule
\cmark & \xmark & \xmark & \xmark & \xmark & \multicolumn{1}{c|}{\xmark} & 66.7                         & 49.9                         & 61.8                         \\
\xmark & \cmark & \xmark & \xmark & \xmark & \multicolumn{1}{c|}{\xmark} & 67.0                         & 50.5                         & 62.2                         \\
\xmark & \xmark & \cmark & \xmark & \xmark & \multicolumn{1}{c|}{\xmark} & 67.3                         & 50.8                         & 62.4                         \\
\xmark & \xmark & \xmark & \cmark & \xmark & \multicolumn{1}{c|}{\xmark} & 67.4                         & 50.9                         & 62.6                         \\
\xmark & \xmark & \xmark & \xmark & \cmark & \multicolumn{1}{c|}{\xmark} & 67.4                         & 50.8                         & 62.5                         \\
\xmark & \xmark & \xmark & \xmark & \xmark & \multicolumn{1}{c|}{\cmark} & 67.7                         & 51.3                         & 63.0                         \\ \midrule
\cmark & \cmark & \cmark & \cmark & \cmark & \multicolumn{1}{c|}{\cmark} & 60.4                         & 43.4                         & 55.8                         \\
\xmark & \cmark & \cmark & \cmark & \cmark & \multicolumn{1}{c|}{\cmark} & 64.1                         & 47.0                         & 59.3                         \\
\xmark & \xmark & \cmark & \cmark & \cmark & \multicolumn{1}{c|}{\cmark} & 65.3                         & 49.2                         & 60.0                         \\
\xmark & \xmark & \xmark & \cmark & \cmark & \multicolumn{1}{c|}{\cmark} & 67.5                         & 50.9                         & 62.7                         \\
\xmark & \xmark & \xmark & \xmark & \cmark & \multicolumn{1}{c|}{\cmark} & \textbf{68.0}                & \textbf{51.4}                & \textbf{63.3}                \\ \bottomrule
\end{tabular}
\end{adjustbox}
\caption{Performance of different text-vision fusion locations in the encoder of our best model (\texttt{VG-BART+FG+VTF} with cross-modal multi-head attention). \cmark \text{ } indicates the occurrence of fusion at a certain layer and \xmark \text{ } indicates non-occurrence. The first row is the result of BART using transcript only.}
\label{tab:fusion_layers_encoder}
\end{table}

\begin{table}[t]
\centering
\begin{adjustbox}{width=\linewidth,totalheight={\textheight},keepaspectratio}
\begin{tabular}{ccccccccc}
\toprule
\multicolumn{6}{c}{\textbf{Decoder Layer \scriptsize{(BART-base)}}}                             & \multirow{2}{*}{\textbf{R-1}} & \multirow{2}{*}{\textbf{R-2}} & \multirow{2}{*}{\textbf{R-L}} \\ \cmidrule{1-6}
1      & 2      & 3      & 4      & 5      & 6                           &                              &                              &                              \\ \midrule
\xmark & \xmark & \xmark & \xmark & \xmark & \multicolumn{1}{c|}{\xmark} & 64.0                & 46.4                & 58.9                \\\midrule
\cmark & \xmark & \xmark & \xmark & \xmark & \multicolumn{1}{c|}{\xmark} & 64.6                         & 47.1                         & 59.6                         \\
\xmark & \cmark & \xmark & \xmark & \xmark & \multicolumn{1}{c|}{\xmark} & \textbf{65.2}                         & \textbf{48.0}                         & \textbf{60.3}                         \\
\xmark & \xmark & \cmark & \xmark & \xmark & \multicolumn{1}{c|}{\xmark} & 64.9                         & 46.9                         & 59.6                         \\
\xmark & \xmark & \xmark & \cmark & \xmark & \multicolumn{1}{c|}{\xmark} & 64.8                         & 46.9                         & 59.7                         \\
\xmark & \xmark & \xmark & \xmark & \cmark & \multicolumn{1}{c|}{\xmark} & 64.3                         & 46.6                         & 59.1                         \\
\xmark & \xmark & \xmark & \xmark & \xmark & \multicolumn{1}{c|}{\cmark} & 64.4                         & 46.7                         & 59.0                         \\
\bottomrule
\end{tabular}
\end{adjustbox}
\caption{Performance of different fusion locations in the decoder of our best model (\texttt{VG-BART+FG+VTF} with cross-modal multi-head attention).}
\label{tab:fusion_layers_decoder}
\end{table}

\subsection{Where to Inject Visual Information} \label{sec:where_to_inject}
As discussed in Section~\ref{sec:intro}, one of the main challenges of building VG GPLMs is to find the optimal location to inject the visual information (i.e., the text-vision fusion). A sub-optimal location might lead to a less effective modality fusion and even hurt the GPLMs' original text generation ability. As GPLMs have a stack of layers in the encoder and also the decoder, we explore this problem from two aspects: 1) which single layer has the best fusion effect; and 2) does multiple times of fusion help GPLMs to understand the visual information better?

As depicted in Table~\ref{tab:fusion_layers_encoder} and \ref{tab:fusion_layers_decoder}, firstly, we enumerate each single layer in the encoder and decoder of our best model (\texttt{VG-BART+FG+VTF}) to perform the text-vision fusion. 
In terms of ROUGE scores, we can clearly tell that injecting visual information into the encoder can generally boost the model's performance by a large step, while injecting into the decoder only shows negligible improvement. 
Furthermore, in the encoder, we observe that injecting at a higher layer (closer to the encoder output) brings more improvement. Instead, in the decoder, there is no clear pattern showing the influence of injecting location.
We speculate that an early text-vision fusion in the encoder makes the visual information slightly fades away after passing through the stack of encoder layers. 
Additionally, during the decoding stage, the model utilizes visual information better through the encoder-decoder attention layers than directly injecting into the decoder, which could potentially hurts the generation ability.
Secondly, as shown in the lower part of Table~\ref{tab:fusion_layers_encoder}, we conduct multiple times of fusion in the encoder's different locations. We observe that when fusing at all encoder layers simultaneously, the model converges to a much worse performance. We conjecture that this causes the catastrophic forgetting of the pre-trained knowledge in GPLMs. We find that fusing at the last several layers (e.g., 5 and 6) in the encoder is able to further improve the summarization performance.

\subsection{Effects of the Forget Gate}\label{sec:effect_of_fg}
\begin{figure}[t]
    \centering
    \includegraphics[width=\linewidth]{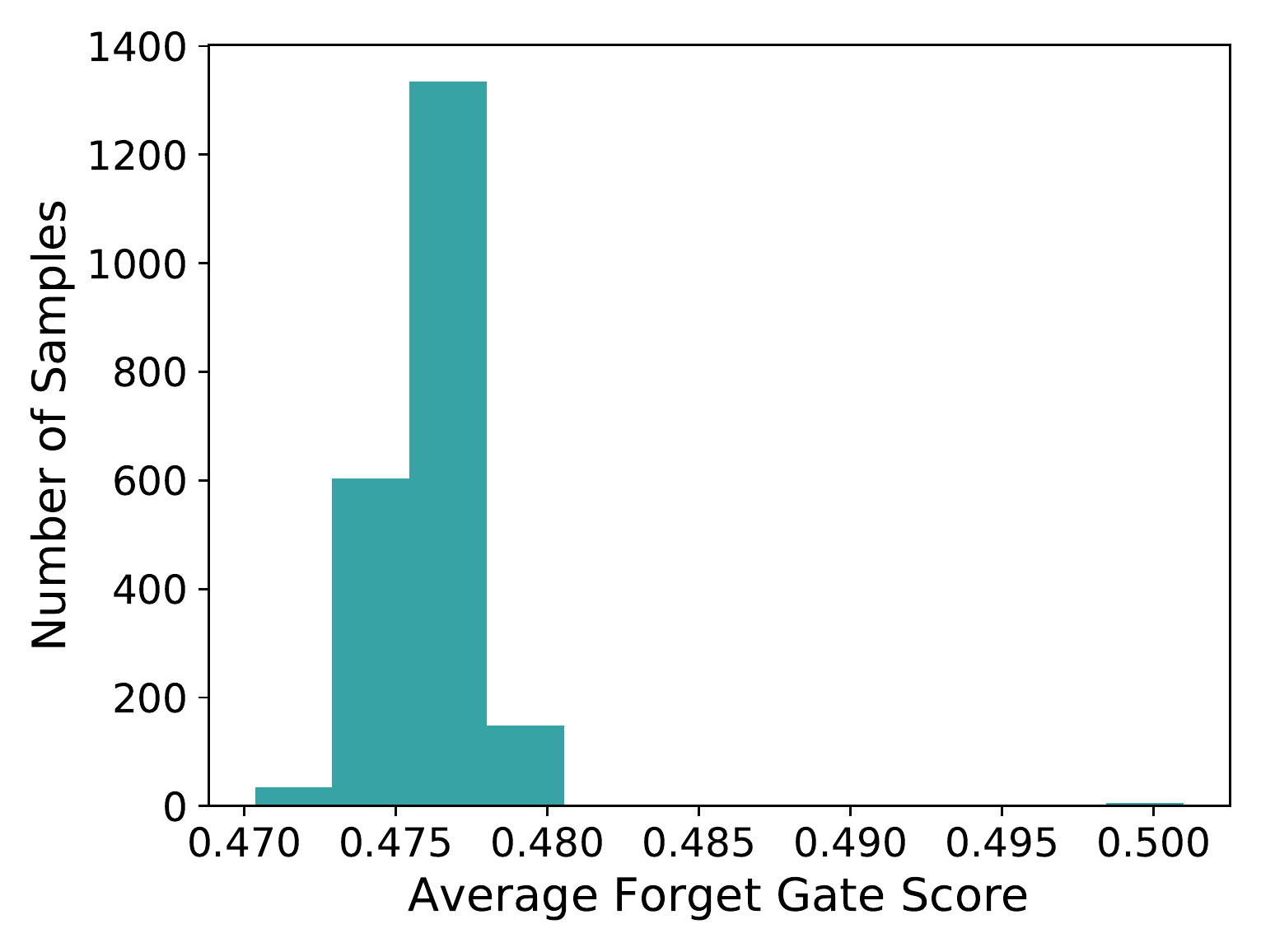}
    \caption{The distribution of average forget gate score on the How2 test set. The model is the VG-BART with dot-product attention.}
    \label{fig:forget_gate}
\end{figure}

As mentioned in Section \ref{sec:fusions}, we apply a forget gate (Eq.\ref{eq:fg_mask}) to filter out noise and let the model focus on more important visual information.
To have a deeper understanding of the effects of the forget gate, 
we calculate the average forget gate score (averaged over the whole sequence) for each sample from the How2 test set. As shown in Figure \ref{fig:forget_gate}, most scores are distributed between 0.47 and 0.48. 
There is one data sample the score reaches 0.5 because its transcript is not available.
As illustrated in Table~\ref{tab:case-study}, the model can still generate reasonable summary for it by paying more attention to the visual information. The meaning of the generated summary is still highly aligned with the reference summary, which shows the capability and flexibility of our model to utilize visual information. 

\begin{table}[t]
\centering
\begin{adjustbox}{width=\linewidth,totalheight={\textheight},keepaspectratio}
\begin{tabular}{p{1\columnwidth}}
\toprule
\textbf{Transcript}: transcript not available

\textbf{Summary from \textit{Transcript + Video}}: learn tips on how to write ``cane'' in chinese radicals with mandarin characters in the free video clip. get free foreign language lessons from an expert.                     \\
\textbf{Reference Summary}: learn what ticks are in chinese calligraphy in this free video clip on languages and writing.                                             \\ \bottomrule
\end{tabular}
\end{adjustbox}
\caption{An example from How2 testing dataset that has high forget gate score.}
\label{tab:case-study}
\end{table}

\section{Conclusion and Future Work}
\label{sec:conclusion}
In this paper, we introduce a simple yet effective method to construct vision guided large-scale generative pre-trained language models (VG-BART and VG-T5) for the multimodal abstractive summarization task by inserting attention-based add-on layers. We propose two types of attention mechanisms for the text-vision fusion and interaction: 1) Cross-modal Dot-product Attention; and 2) Cross-modal Multi-head Attention. Moreover, we also investigate the effects of using the forget gate and visual transformer encoder along with the attention mechanisms. In addition, we enumerate almost all possible locations in GPLMs for injecting add-on layers. Experimental results show that our approaches significantly outperform the prior state-of-the-art on the How2 dataset.  Further analysis illustrates that multi-head attention is more robust than the dot-product attention and higher layers of the encoder is the optimal place to inject vision information. For future work, we believe that our analyses on the \textit{how} and \textit{where} to inject visual information into GPLMs can be applied to other multimodal tasks.

\section{Acknowledgments}
We want to thank the anonymous reviewers for their constructive feedback. This work is partially funded by ITS/353/19FP and and MRP/055/18 of the Innovation Technology Commission, the Hong Kong SAR Government.

\bibliography{anthology,custom}
\bibliographystyle{acl_natbib}

\appendix

\section{Multi-head Self-Attention} \label{appendix:msa}
The query ($Q$), key ($K$), value ($V$) based self-attention is the core building block of the Transformer model~\citep{vaswani2017attention}. Given the input $Z \in \mathbb{R}^{s \times d_z}$, we calculate $Q$, $K$, and $V$ by
\begin{align*}
    Q &= Z W_q, Q \in \mathbb{R}^{s \times d_k} \\
    K &= Z W_k, K \in \mathbb{R}^{s \times d_k} \\
    V &= Z W_v, V \in \mathbb{R}^{s \times d_v},
\end{align*}
in which $W_q \in \mathbb{R}^{d_z \times d_k}$, $W_k \in \mathbb{R}^{d_z \times d_k}$, and $W_v \in \mathbb{R}^{d_z \times d_v}$ are the projection weights. Then, a single-head self-attention is calculated by
\begin{align*}
    \text{Attention}(Q, K, V) = \text{softmax}(\frac{QK^T}{\sqrt{d_k}})V,
\end{align*}
where $\frac{1}{\sqrt{d_k}}$ is the scaling factor to mitigate the extremely small gradients issue mentioned by \citet{vaswani2017attention}. For multi-head self-attention, it can be calculated by 
\begin{align*}
    \text{MultiHead}(Q, K, V) = \text{Concat}(\text{head}_1, ..., \text{head}_h)W_o
\end{align*}
and 
\begin{align*}
    \text{head}_i = \text{Attention}(QW_q^i, KW_k^i, VW_v^i).
\end{align*}

\section{Feed-Forward Network} \label{appendix:ffn}
Given the input $Z \in \mathbb{R}^{s \times d_z}$, the feed-forward network (FFN) processes it with two linear projections $W_f^1 \in \mathbb{R}^{d_z \times d_f}$, $W_f^2 \in \mathbb{R}^{d_f \times d_z}$ and a non-linear function GELUs~\citep{Hendrycks2016GaussianEL},
\begin{align*}
    \text{FFN}(Z) = \text{GELU}(Z W_f^1) W_f^2.
\end{align*}
In addition, after each linear projection, there is a dropout~\citep{Srivastava2014DropoutAS} layer to improve generalization.

\end{document}